\pdfoutput=1

\documentclass[11pt]{article}

\usepackage[]{EMNLP2023}
\usepackage{times}
\usepackage{url}
\usepackage{latexsym}
\usepackage{multirow}
\usepackage{colortbl}
\usepackage{booktabs}
\usepackage{amssymb}
\usepackage{amsmath}
\usepackage{amsthm}
\usepackage{graphicx}
\usepackage{makecell}
\usepackage{threeparttable}
\usepackage{CJKutf8}
\usepackage{fancyhdr}
\usepackage{subfigure}
\usepackage{float}
\usepackage{supertabular}
\usepackage{tablefootnote}
\usepackage{booktabs}
\usepackage{subfigure}
\usepackage{algorithm}
\usepackage{algorithmic}
\usepackage{color}
\usepackage{microtype}
\usepackage{amsmath, amssymb, amscd, amsfonts}
\usepackage{IEEEtrantools}
\usepackage{soul}
\usepackage{url}
\usepackage{algorithm}
\usepackage{algorithmic}
\usepackage{amssymb}
\usepackage{amsfonts} 
\usepackage{multirow}
\usepackage{boldline}
\usepackage{hhline}
\usepackage{natbib} 
\usepackage{amssymb}
\usepackage{amsfonts} 
\usepackage{multirow}
\usepackage{boldline}
\usepackage{hhline}
\usepackage{amsmath}
\usepackage{xcolor}
\usepackage{pifont}
\usepackage{subfigure}
\usepackage{makecell}
\usepackage{mathrsfs}
\usepackage{utfsym}
\usepackage[normalem]{ulem}
\useunder{\uline}{\ul}{}
\usepackage[inline]{enumitem}
\usepackage{enumitem}
\usepackage{latexsym}

\usepackage[T1]{fontenc}

\usepackage[utf8]{inputenc}

\usepackage{microtype}

\usepackage{inconsolata}

\usepackage{CJKutf8}
\usepackage{makecell}

\title{\texttt{EduChat}: A Large-Scale Language Model-based Chatbot System \\for Intelligent Education}

\author{Yuhao Dan$^{1*}$,~Zhikai Lei$^{1*}$,~Yiyang Gu$^1$\thanks{$^*$ Equal contribution.},~Yong Li$^1$,~Jianghao Yin$^1$,\\
\textbf{Jiaju Lin}$^1$\textbf{,~Linhao Ye}$^1$\textbf{,~Zhiyan Tie}$^1$\textbf{,~Yougen Zhou}$^1$\textbf{,~Yilei Wang}$^2$\textbf{,~Aimin Zhou}$^{1,2}$\textbf{,}\\
\textbf{~Ze Zhou}$^{4}$ \textbf{~Qin Chen}$^{1\dag}$ \textbf{,~Jie Zhou}$^{1}$\thanks{$^\dag$ Corresponding author.} \textbf{,~Liang He}$^1$ 
\textbf{,~Xipeng Qiu}$^3$\\
{$^1$ \normalsize School of Computer Science and Technology, East China Normal University, Shanghai, China} \\
 {$^2$ \normalsize Institute of AI for Education, ECNU, Shanghai, China} \\
{$^3$ \normalsize School of Computer Science, Fudan University, Shanghai, China} \\
{$^4$ \normalsize ZhuQingTing Data Technology (Zhejiang) Co., Ltd., Zhejiang, China} \\
   }
   
\begin{document}
\maketitle
\begin{abstract}
\texttt{EduChat}\footnote{https://www.educhat.top/} is a large-scale language model (LLM)-based chatbot system in the education domain. Its goal is to support personalized, fair, and compassionate intelligent education, serving teachers, students, and parents. Guided by theories from psychology and education, it further strengthens educational functions such as open question answering, essay assessment, Socratic teaching, and emotional support based on the existing basic LLMs. 
Particularly, we learn domain-specific knowledge by pre-training on the educational corpus and stimulate various skills with tool use by fine-tuning on designed system prompts and instructions. 
Currently, \texttt{EduChat} is available online as an open-source project, with its code, data, and model parameters available on platforms (e.g., GitHub\footnote{https://github.com/icalk-nlp/EduChat}, Hugging Face\footnote{https://huggingface.co/ecnu-icalk}). We also prepare a demonstration of its capabilities online\footnote{https://vimeo.com/851004454?share=copy}.
This initiative aims to promote research and applications of LLMs for intelligent education.
\end{abstract}

\section{Introduction}
Recently, large-scale language models (LLMs), such as ChatGPT \cite{schulman2022chatgpt}, LLaMa \cite{touvron2023llama}, have achieved great success in the field of natural language processing \cite{zhou2023chatgpt}. 
LLMs obtained the ability of reasoning, long-range context modeling, and task generalization by training on large-scale textual corpus with some strategies, such as code pre-training~\citep{chen2021evaluating}, instruction tuning~\citep{DBLP:conf/iclr/WeiBZGYLDDL22}, and reinforcement learning from human feedback (RLHF)~\citep{stiennon2020learning}.
With the advent of LLMs, they have the potential to revolutionize intelligent education by providing personalized, comprehensive, and timely support to teachers, students, and parents. 

However, there are several challenges of applying LLMs into education domain.
One challenge (\textbf{C1}) is that there is still a gap between the LLMs and the educational expert since LLMs are pre-trained on the general corpus, which lack sufficient educational knowledge and can not align well with real scenarios (e.g., essay assessment).
The other challenge (\textbf{C2}) is that the knowledge in the field of education is updating, while LLMs can not learn up-to-date knowledge due to the training mechanism. Moreover, LLMs suffer from the hallucination problem, and may generate responses that are not truthful.

To address these problems, we propose \texttt{EduChat}, an LLM-based chatbot system for intelligent education. 
For \textbf{C1}, we pre-train LLMs on a large number of educational books (e.g., psychology, ancient poetry) and 4 million cleaned diverse instructions to learn the fundamental knowledge. Then, we fine-tune the model on 500 thousand high-quality customized instructions to activate education-specific functions (e.g., essay assessment, Socratic teaching and emotional support), by aligning with the feedbacks from psychology experts and frontline teachers. 
For \textbf{C2}, we explore a retrieval-augmented technology, which enables LLMs to automatically judge the helpfulness of the retrieved information, and generate the response based on the relevant information and knowledge stored in LLMs.
In this way, our \texttt{EduChat} can access the latest information from the internet, ensuring that the responses are accurate and credible.
As an open-source project, \texttt{EduChat} improves the performance of education-specific functions while maintaining comparable foundational capabilities to other large-scale models with equivalent parameter size. The main contributions are as follows:
\begin{itemize}[leftmargin=*, align=left]
\vspace{-1mm}
    \item We explore the potential of incorporating theories of psychology and education into LLMs, which sheds light on how to adapt general LLMs to specific domains;
    \vspace{-1mm}
    \item Diverse system prompts and instructions are designed to control the tool use and stimulate different skills, which alleviates the problem of hallucination and is more applicable in real education scenarios;
    \vspace{-1mm}
    \item We develop and release the \texttt{EduChat} system with various educational functions, thus developers and researchers can help speed up the research and applications of intelligent education.
\end{itemize}

\section{Related Work}
\vspace{-1mm}
Recently, LLMs like ChatGPT \cite{schulman2022chatgpt}, ChatGLM \cite{du2022glm}, and LLaMA2-Chat \cite{Touvron2023Llama2O} have emerged as a breakthrough technology in natural language processing, achieving strong performance on language generation and understanding through pre-training on massive text and instruction tuning.

While LLMs demonstrate impressive capabilities in general domains, their lack of subject-matter expertise becomes apparent when applied to specialized verticals.
For instance, we can find specialized language models catering to various domains, such as ChatDoctor \cite{li2023chatdoctor} and HuaTuoGPT \cite{Zhang2023HuatuoGPTTT} in healthcare, FinGPT \cite{Yang2023FinGPTOF} in finance, and ChatLaw \cite{Cui2023ChatLawOL} in the legal domain.
These niche fields inherently necessitate models to possess comprehensive domain knowledge to address relevant queries, especially when assisting real users in practical scenarios. 
In education, \citet{baladn-etal-2023-retuyt} tune open-source LLMs for generating better teacher responses in BEA 2023 Shared Task \cite{tack-etal-2023-bea}. But challenges still exist, such as the lack of domain knowledge in general LLMs and the necessity for them to align with educational abilities (e.g., essay assessment, emotional support, and Socratic teaching). \texttt{EduChat} is pre-trained on a diverse education corpus to ensure the alignment of \texttt{EduChat} with educational abilities.

\vspace{-1mm}
\section{Core Functions of \texttt{EduChat}}
\vspace{-1mm}
\paragraph{Retrieval-Augmented Open Question Answering (QA)}
The education domain demands high accuracy and real-time updates regarding knowledge and related policies. However, existing generative LLMs suffer from issues like fabricating information and lagging behind in knowledge updates. To address this, we explore retrieval-augmented open QA methods. By utilizing real-time updated corpora from the internet as an external knowledge source, we enable LLMs to autonomously assess the relevance of retrieved information to answer a given question and decide which information to incorporate for generating responses. Through extensive experimental analysis, we discover that our model exhibits significant advantages over general LLMs in terms of eliminating fabrications and maintaining up-to-date knowledge.

\paragraph{Fine-grained Essay Assessment}
In essay assessment, teachers meticulously annotate grammar errors, provide scores, and offer feedback on standout sentences. Existing language models often have coarse granularity in grading, limiting students' writing skill improvement. Our research focuses on more fine-grained and comprehensive essay assessment. Combining frontline teaching professionals' expertise, we provide overall scores, aspect-level ratings, and detailed comments on content, expression, paragraph, and overall evaluation. Our model can identify standout sentences, highlighting strengths and areas for improvement, enabling personalized guidance for students' essay writing skills. This ensures timely and professional support in all aspects of writing.

\paragraph{Socratic Teaching}
We focus on developing Socratic teaching capabilities in LLMs rather than providing direct answers to students. 
We adopt the Socratic dialogue method, engaging in multi-step question-and-answer interactions to encourage independent thinking. By stimulating discussions, debates, evaluations, and analyses, we aim to foster advanced cognitive skills and cultivate students' autonomy in learning. Our ultimate goal is to enhance critical thinking and innovation abilities to their fullest extent.

\paragraph{Psychology-based Emotional Support}
Adolescents and children face more severe psychological pressures due to their immature cognitive development. Whereas, current LLMs usually provide generic advice, which can not well fit the specific emotional problem. To address this, we develop a psychological inquiry framework based on emotion psychology, such as Rational Emotive Behavior Therapy (REBT) and the ABC theory \cite{ellis_revised_1991}. Our fine-tuned model can simulate a psychological counselor, providing personalized diagnoses and emotional support for users. \texttt{EduChat} fosters a deeper understanding of users' emotional states and offers accurate and professional assistance.


\section{Data Construction}
\subsection{Pre-training Data}
\label{pretraining_data}

\paragraph{Textbooks Data}
In our research, we gather a vast amount of educational textbook and online question bank data from Chinese middle and high school exams for pre-training. Additionally, we enrich our model with over 70,000 Chinese poetries, providing detailed information on authors, backgrounds, and poetry appreciation to enhance its poetry creation and appreciation capabilities. To facilitate empathetic emotional support dialogues, we carefully select 60 famous works from hundreds of psychology books. These selected books belong to two main categories. The first category consists of 15 branches of psychological theory, including developmental and educational psychology, social psychology, behavioral psychology, counseling psychology and others. The second category contains various psychological practices, which offer practical cases of psychological consultation and emotional support dialogues. By incorporating the diverse fundamental data into pre-training, our model gains a deeper understanding of education and psychology, enabling it to generate more helpful responses.

\paragraph{Fundamental Instruction Data}
To achieve a more natural human-computer interaction, we collect a large volume of bilingual instruct tuning data from reputable open-source repositories like Alpaca\footnote{https://github.com/tatsu-lab/stanford\_alpaca}, BELLE \cite{belle2023exploring}, GPT4All\footnote{https://github.com/nomic-ai/gpt4all}, Open-Assistant\footnote{https://github.com/LAION-AI/Open-Assistant}, FLANCoT\footnote{https://huggingface.co/datasets/lucasmccabe-lmi/FLAN\_CoT\_alpaca\_style}, and Firefly\footnote{https://github.com/yangjianxin1/Firefly}. The data spans various task types, enabling our models to acquire foundational instruction following capabilities for diverse instruction types. In addition, we source high-quality multi-turn dialogue data from MOSS \cite{sun2023moss}, BELLE \cite{belle2023exploring}, COIG \cite{zhang2023chinese}, LIMA \cite{zhou2023lima}, and ShareGPT\footnote{https://huggingface.co/datasets/gozfarb/ShareGPT\_\\Vicuna\_unfiltered}. This data covers various dialogue contexts, including role-playing, creative writing, and code-related discussions, ensuring our models' competence in engaging and sustaining meaningful multi-turn conversations.



\begin{figure}[t!]
    \centering
    \includegraphics[scale=0.35]{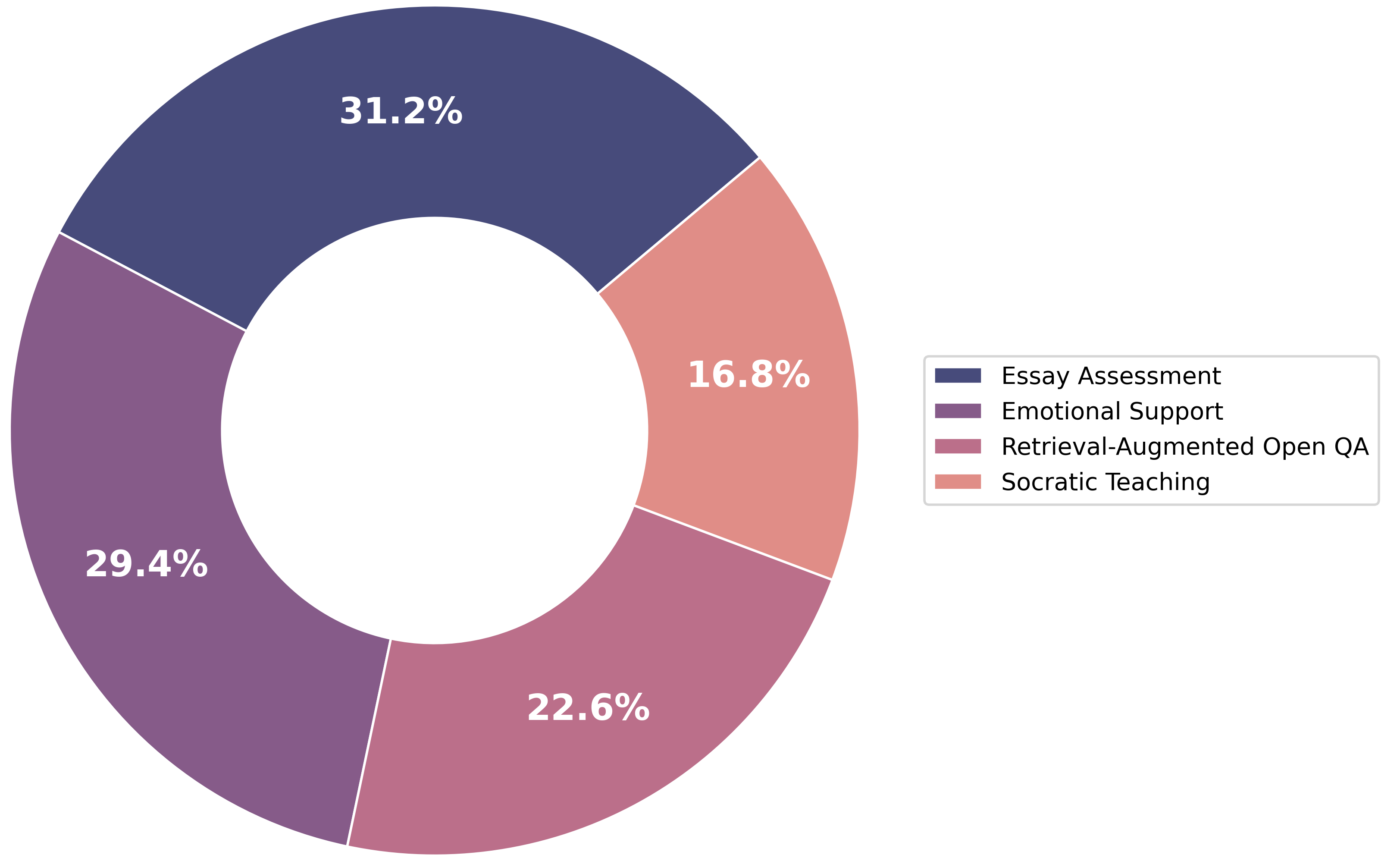}
    \vspace{-1mm}
    \caption{Distribution of educational data.}
    \label{fig:data_dis}
\vspace{-4mm}
\end{figure}

\subsection{Fine-tuning Data}
\label{sft_data}
\vspace{-1mm}

To enhance the capability of education, we construct the \textbf{Educational Instruction Data} for fine-tuning, which covers retrieval-augmented open QA, emotional support, Socratic teaching and essay assessment. The distribution is shown in Figure \ref{fig:data_dis}.

\paragraph{Retrieval-Augmented Open QA Data}
To address hallucination and timely knowledge issues in Open QA, we design a retrieval-augmented open QA technique. We sample high-quality data through ChatGPT scoring in relevant Open QA and Subject QA datasets. To tackle irrelevant retrieved content, we introduce self-checking. ChatGPT assesses whether the retrieval content helps answer the question and then generates the answer using an self-check, incorporating the useful retrieval content and questions. To maintain data quality, we manually verify the data during this process.

\begin{figure*}[!t]
\begin{center}
\includegraphics[width=0.9\textwidth]{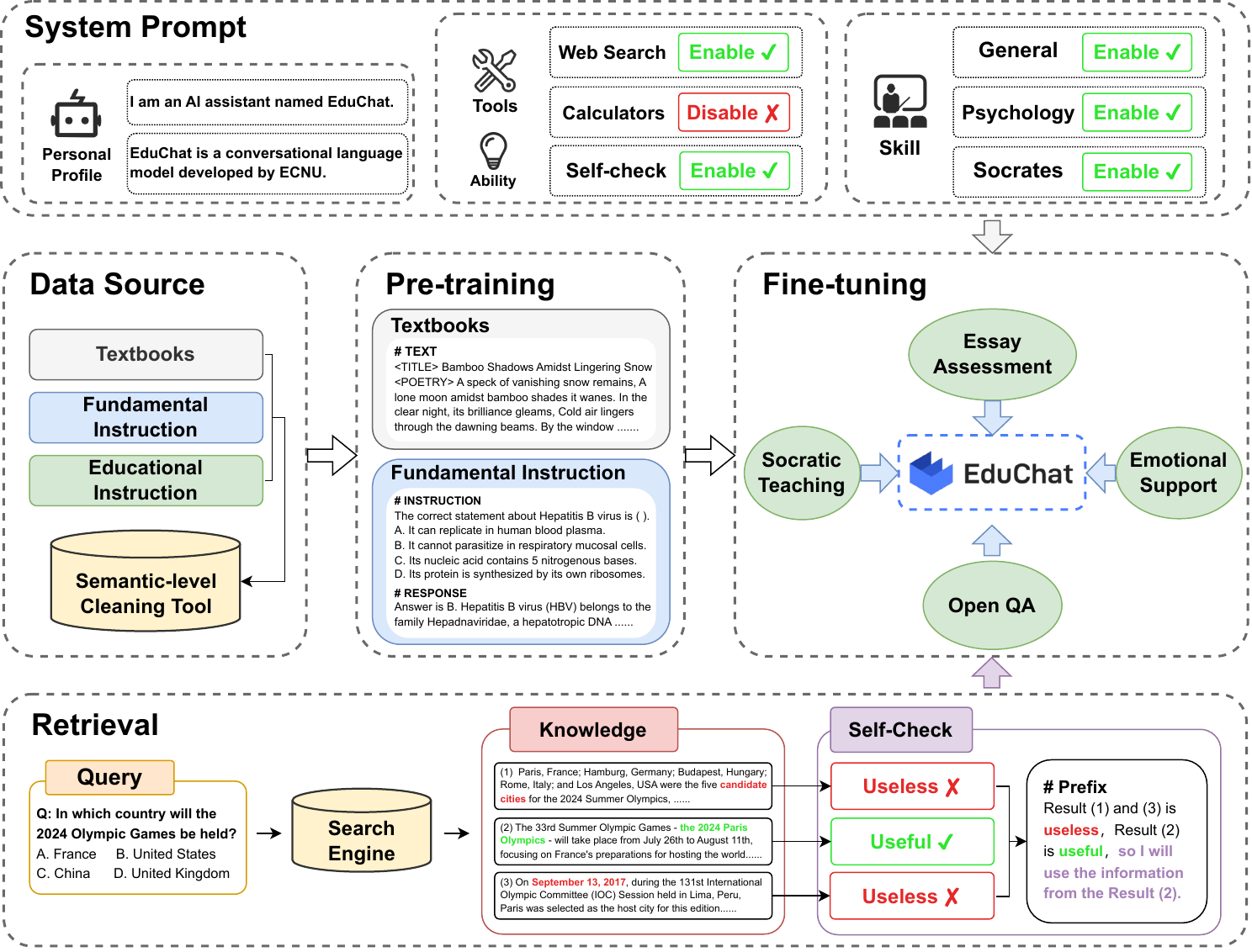}
\end{center}
\vspace{-3mm}
\caption{The overall framework of \texttt{EduChat}.} 
\vspace{-4mm}
\label{fig:educhat_framework}
\end{figure*}

\vspace{-1mm}
\paragraph{Emotional Support Data}
To overcome the scarcity of Chinese emotional support dialogue data, we adopt a translation and expansion approach. We translate the widely-used English emotional support dataset, ESConv \cite{liu-etal-2021-towards}, into Chinese as ESConv-zh. 
After manual review and cleaning, we simulate multi-agent dialogues based on various patient scenarios within ESConv-zh and also collect real-life Chinese psychological counseling consultation data, incorporating patient information and diagnosis results.
By training our models on diverse datasets, we empower them to provide robust emotional support and act as compassionate counselors during consultations.

\vspace{-1mm}
\paragraph{Socratic Teaching Data}

Teachers play a key role in guiding and encouraging heuristic exploration rather than just providing answers. 
To support this, we generate dialogues simulating the Socratic teaching method by incorporating multi-step Q\&A involving counter-questions, challenges, and inquiries.  
These dialogues are manually evaluated for accuracy, fluency, and progression from easy to complex questions. 
Integrating this dataset into training equips our model with a strong capability in Socratic teaching, distinguishing it from other LLMs that only offer direct answers.

\vspace{-1mm}
\paragraph{Essay Assessment Data}
The lack of timely and detailed feedback often hinders students' writing improvement. To tackle this issue, we create a high-quality essay assessment dataset. Initially, we collect essays and employ ChatGPT to evaluate them in terms of content, expression, and overall quality. To ensure data quality, we invite pedagogical experts to manually curate the comments. This dataset empowers \texttt{EduChat} with the ability to provide students with high-quality feedback, aiding in the enhancement of their writing skills.

\subsection{Data Preprocessing}
To enhance data quality, we conduct semantic-level deduplication on the dataset. Using the sentence-transformers model \citep{sentence-bert}, we obtain sentence embeddings for each data point and calculate cosine similarity between all pairs of embeddings. For similarities exceeding a threshold of 0.7, we remove one of the duplicates. 
We implement the similarity calculation using CUDA for GPU acceleration, speeding up the process.

\begin{figure*}[t!]
\vspace{-2mm}
    \centering
    \includegraphics[scale=0.28]{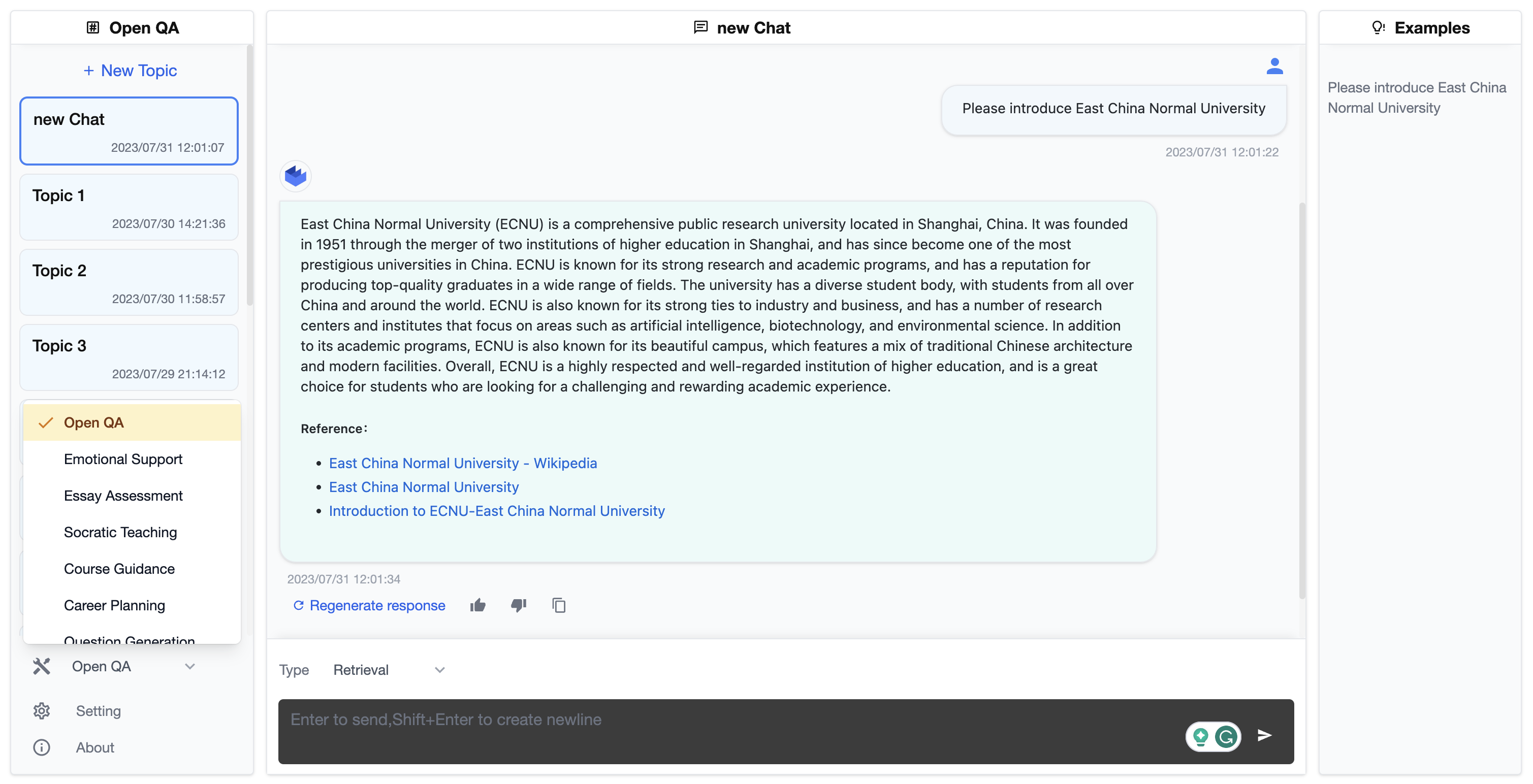}
    \vspace{-3mm}
    \caption{Demo of \texttt{EduChat}.}
    \label{fig:demo}
\vspace{-3mm}
\end{figure*}

\section{\texttt{EduChat}}

\texttt{EduChat} is an LLM designed for the education domain (Figure \ref{fig:educhat_framework}).
We first \textbf{pre-train} on large-scale education corpus (e.g., textbooks, instructions for foundational tasks) to learn the domain-specific and foundational knowledge. 
We then learn the pedagogical skills by \textbf{fine-tuning} \texttt{EduChat} on task-specific instruction datasets. 
Moreover, we leverage online \textbf{retrieval} to enhance the accuracy and timeliness of knowledge in its responses. 
To control skills, we design various \textbf{system prompts} to unlock different scenes with tool usage. 

\subsection{Training Procedure of \texttt{EduChat}}
The training of \texttt{EduChat} is mainly divided into two stages: fundamental capabilities acquisition and educational skills acquisition. 
In the first stage, we \textbf{pre-train} the model on educational books and Q\&A pairs (detailed in Section \ref{pretraining_data}) to equip it with foundational knowledge across disciplines. Besides, large-scale instruction tuning and open-domain dialogue datasets are also incorporated to enable basic instruction following ability and dialogue ability (detailed in Section \ref{sft_data}). 
In the second stage, we develop \texttt{EduChat}'s pedagogical skills by \textbf{fine-tuning} the model on our carefully curated data, including retrieval-augmented open QA, emotional support, Socratic teaching and essay assessment datasets mentioned in Section \ref{sft_data}. 

\subsection{Online Knowledge Retrieval}
Existing generative LLMs all suffer from the issues of generating hallucinations and outdated information, which is detrimental to an educational model. To mitigate this problem, we introduce self-check as shown in Figure \ref{fig:educhat_framework}. Specifically, when online knowledge retrieval is enabled, the model picks useful retrieval results by asking itself "Is this helpful for answering the question?" and append filtered snippets before the dialogue history.

\subsection{System Prompt Design}
Teachers always utilize various tools with different skills to enhance their teaching across different subjects. To enable \texttt{EduChat} to emulate an authentic teacher-student interaction, we carefully craft the system prompt that consists of personal profile, tool usage and skill selection (see Figure \ref{tab:sys_prompt}). Detailed settings can be found in Table \ref{tab:sys_prompt}.

\textbf{1) Personal Profile:} To remind the model of its own identity, the system prompt begins with: "\texttt{EduChat} is a conversational language model developed by East China Normal University."; 
\textbf{2) Tool Usage:} To regulate tool availability, the second part of the system prompt commences with "\texttt{EduChat}'s tools:", listing all tool names and their respective accessibility. For instance, "Web search: Enable" indicates the model's ability to use retrieval, while "Calculator: Disable" signifies the model's inability to utilize a calculator; 
\textbf{3) Skill Selection:} 
Teachers in various settings possess unique communication skills, such as Socratic Teaching or Psychology-based Emotional Support. 
To cater to specific scenarios, we include function names at the end of the system prompt, which activates corresponding abilities based on the scene's requirements.

\subsection{Demonstration}
We also develop a user-friendly demo system for \texttt{EduChat} (see Figure \ref{fig:demo}). Upon logging in, users can select from various functions, such as Open QA and Emotional Support, each offering a scene-specific system prompt to activate the corresponding ability. 
With this intuitive interface, users can easily engage in interactive conversations with \texttt{EduChat} to assist students, teachers and parents. 
Additionally, the system is designed to be adaptive, continuously learning from user interactions to further improve its capabilities and provide more personalized and effective assistance over time. 

\begin{figure*}[!t]
\begin{center}
\includegraphics[width=1.0\textwidth]{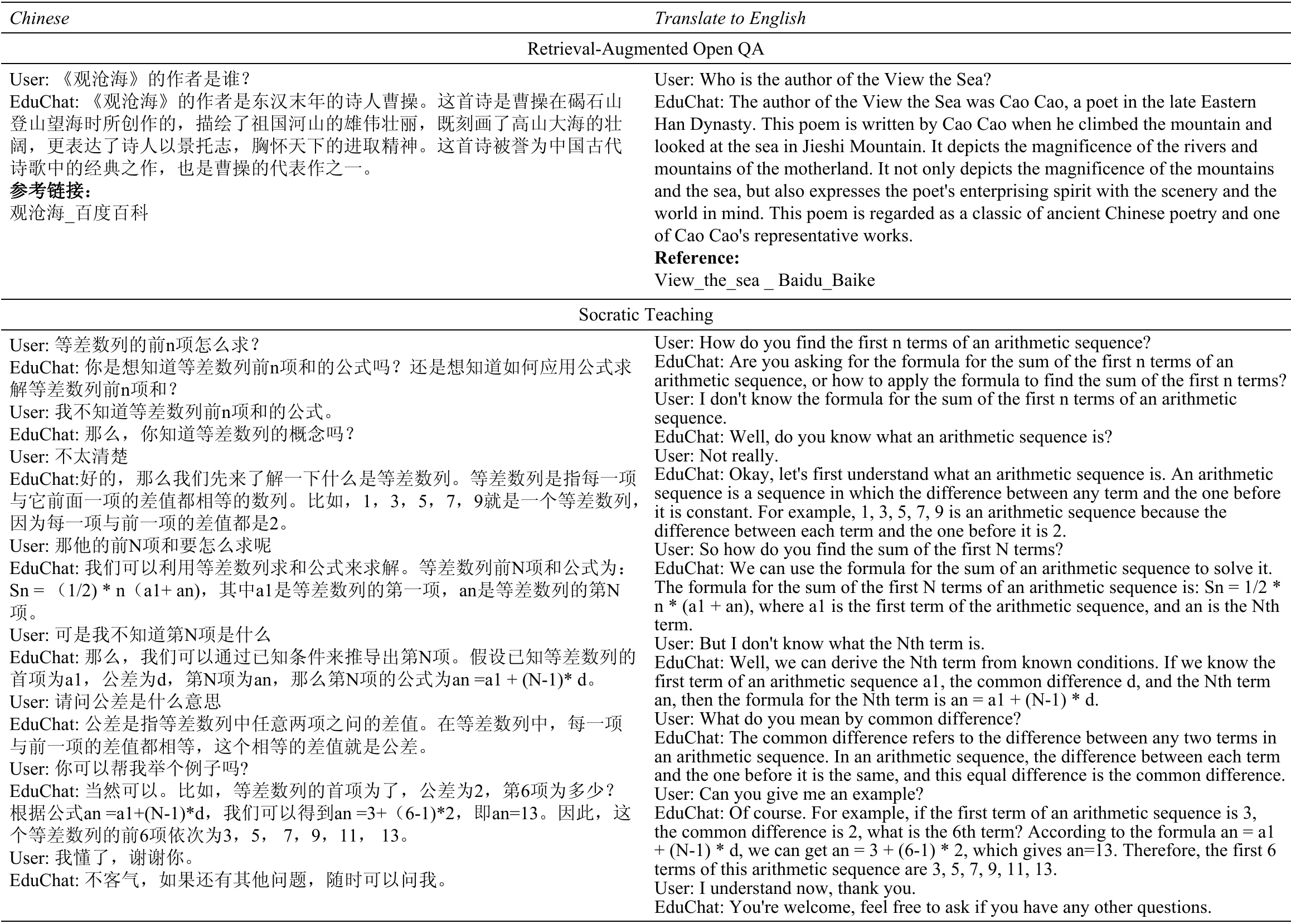}
\end{center}
\vspace{-3mm}
\caption{Cases of retrieval-augmented open QA and socratic teaching.} 
\vspace{-3mm}
\label{fig:case studies 1}
\end{figure*}

\section{Experimental Results}
\subsection{Resutls of C-Eval}
Table \ref{table: c-evel} presents the results of our model on the C-Eval benchmark \cite{huang2023ceval}, a comprehensive Chinese evaluation suite for foundation models. The dataset consists of 13,948 multi-choice questions, spanning 52 diverse disciplines and categorized into four difficulty levels.
Analyzing the table, we observe that our model achieves commendable performance compared to models with similar parameter scales, such as Chinese Alpaca-13B and WastlackLM. Notably, both \texttt{EduChat} and Chinese Alpaca-13B are built on the LLaMa-13B base model. However, \texttt{EduChat} outperforms Chinese Alpaca-13B by over seven points.
Furthermore, our integration of retrieval into LLMs proves to be highly effective, demonstrating the power of our retrieval-augmented open QA technique in enhancing model performance.

\begin{table}[t!]
\centering
\scriptsize
\setlength{\tabcolsep}{0.1mm}{
\begin{tabular}{lcccccc}
  \hlineB{4}
             & \textbf{STEM}            & \textbf{Social Science}  & \textbf{Humanities}      & \textbf{Others}          & \textbf{Avg(hard)}       & \textbf{Avg}             \\ \hline
GPT-4                        & 67.1                     & 77.6                     & 64.5                     & 67.8                & 54.9              & 68.7                 \\
ChatGPT                      & 52.9                     & 61.8                     & 50.9                     & 53.6           & 41.4                & 54.4                    \\
Baichuan-13B                   & 47.0                       & 66.8                     & 57.3                     & 49.8               & 36.7           & 53.6                   \\
InternLM-7B                   & 48.0                       & 67.4                     & 55.4                     & 45.8                            & 37.1       & 52.8           \\
ChatGLM2-6B                  & 48.6                     & 60.5                     & 51.3                     & 49.8            & 37.1               & 51.7                    \\
WestlakeLM-19B                  & 41.6	& 51.0	& 44.3 &	44.5          & 34.9               & 44.6                   \\
Baichuan-7B                  & 38.2	& 52.0	& 46.2	& 39.3    & 31.5	 & 42.8	        \\ 
Chinese-Alpaca-33B                  & 37.0 &	51.6 &  42.3	&  40.3     & 30.3	  &  41.6	        \\ 
Chinese-Alpaca-13B                  & 31.6	& 37.2	& 33.6	& 32.8  & 27.3     &  33.3	     \\ 
\hline
\cellcolor[HTML]{EFEFEF}{\texttt{EduChat}}      & \cellcolor[HTML]{EFEFEF}{36.2} & \cellcolor[HTML]{EFEFEF}{50.7} & \cellcolor[HTML]{EFEFEF}{42.9} & \cellcolor[HTML]{EFEFEF}{37.7} & \cellcolor[HTML]{EFEFEF}{28.3}  & \cellcolor[HTML]{EFEFEF}{40.7} \\
\cellcolor[HTML]{EFEFEF}{\texttt{EduChat} (w Retrieval)}      & \cellcolor[HTML]{EFEFEF}{43.5} & \cellcolor[HTML]{EFEFEF}{59.3} & \cellcolor[HTML]{EFEFEF}{53.7} & \cellcolor[HTML]{EFEFEF}{46.6} & \cellcolor[HTML]{EFEFEF}{33.1} & \cellcolor[HTML]{EFEFEF}{49.3} \\
\hlineB{4}
\end{tabular}}
\vspace{-1mm}
\caption{Results of C-Eval.}
\label{table: c-evel}
\vspace{-3mm}
\end{table}





\subsection{Case Studies}
Figure \ref{fig:case studies 1} shows the cases of our EduChat on retrieval-augmented open QA and socratic teaching. \texttt{EduChat} can provide precise answer with retrieved relevant information, and learn to guide the student to solve the problems like a teacher step by step. 
For emotional support, \texttt{EduChat} can interact like a psychological counselor rather than giving the general advice.
For space limitation, we provide more cases of psychology-based emotional support and fine-grained essay assessment in the Appendix (Figure \ref{fig:case studies 2}). 

\section{Conclusion}
In this paper, we introduce \texttt{EduChat}, an LLM-based chatbot system for intelligent education. Our goal is to provide personalized, fair, and compassionate support to teachers, students, and parents. 
By leveraging psychology and education theories, we enhance educational functions like open QA, essay assessment, Socratic teaching, and emotional support. 
Through pre-training on educational corpus and fine-tuning with task-specific instructions, \texttt{EduChat} demonstrates great performance on the C-Eval benchmark. 
Overall, \texttt{EduChat} exhibits great potential towards revolutionizing intelligent education. 
In future work, we aim to expand \texttt{EduChat} on more functions, such as career planning, course guidance, question generation and so on.



\bibliography{custom}
\bibliographystyle{acl_natbib}

\appendix

\begin{table*}[]
\centering
\small
\setlength{\tabcolsep}{0.8mm}{
\begin{tabular}{l|cc|ccc}
\toprule 
   &  \multicolumn{2}{c|}{\textbf{Tool}} & \multicolumn{3}{c}{\textbf{Skill}} \\ 
     & Retrieval & Self-check & General & Psychology & Socrates \\ \hline
Retrieval-Augmented Open QA   & True         & True          & True       & False          & False       \\
Fine-grained Essay Assessment   & True/False       & True/False        & True       & False          & False        \\
Psychology-based Emotional Support   & False         & False          & False       & True          & False        \\
Socratic Teaching   & False         & False          & False       & False          & True    \\ \bottomrule  
\end{tabular}
}
\caption{Illustration of system prompt design. }
\label{tab:sys_prompt}
\end{table*}

\begin{figure*}[!t]
\begin{center}
\includegraphics[width=1.0\textwidth]{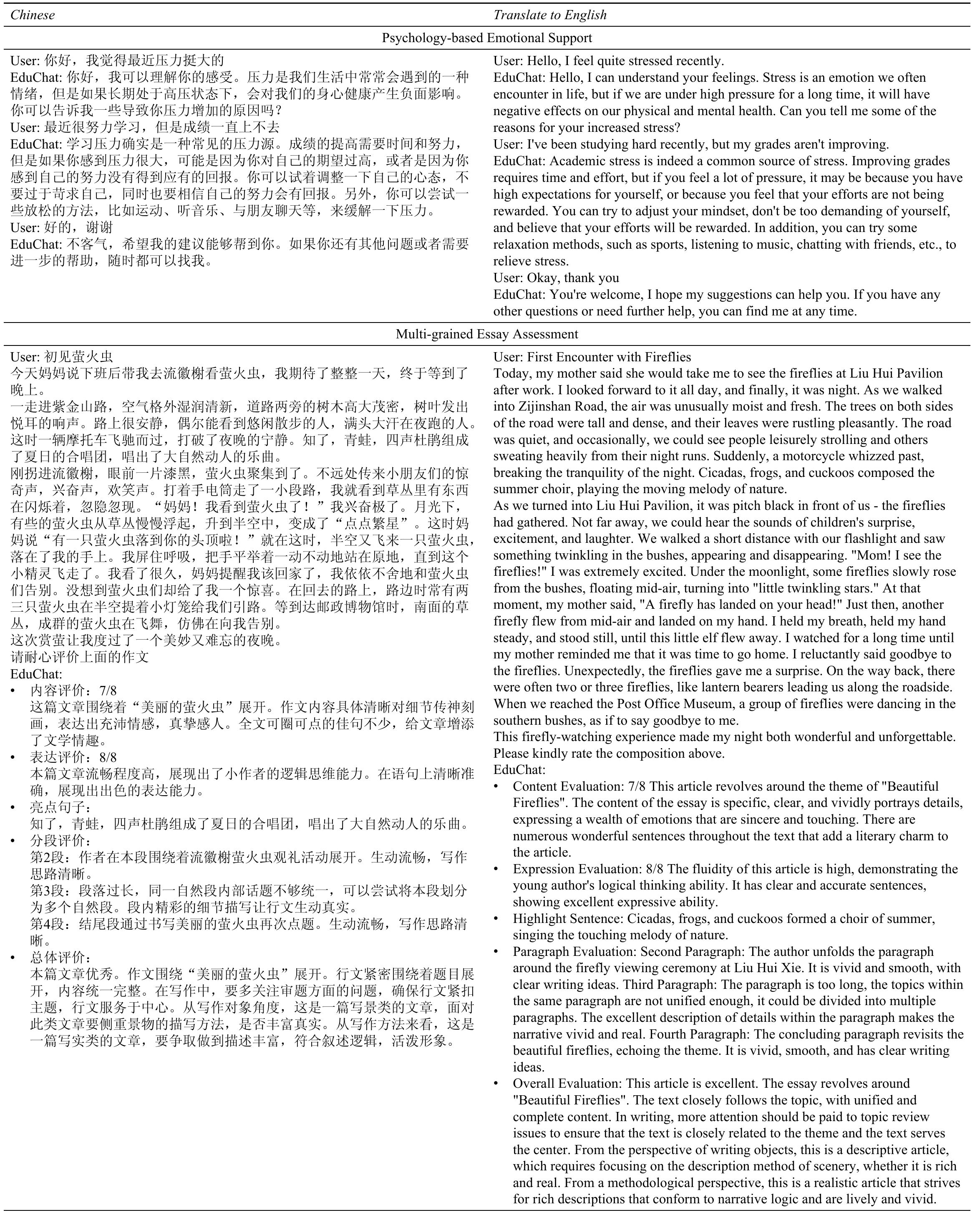}
\end{center}
\caption{Cases of psychology-based emotional support and fine-grained essay assessment.} 
\label{fig:case studies 2}
\end{figure*}

\end{document}